**Machine learning using longitudinal prescription and medical claims for the detection of nonalcoholic steatohepatitis (NASH)**

**Authors** Ozge Yasar[1]*, Patrick Long[2]*, Brett Harder[2], Hanna Marshall[2], Sanjay Bhasin[2], Suyin Lee[2], Mark Delegge[3], Stephanie Roy[2], Orla Doyle[1], Nadea Leavitt[2], John Rigg[1]

**Author Affiliations**

[1.] Real World Solutions, IQVIA, London, United Kingdom, N1 9JY

[2.] Real World Solutions, IQVIA, Plymouth Meeting, Pennsylvania, United States, 19462

[3.] Therapeutic Center of Excellence, IQVIA, Durham, North Carolina, United States, 27703

*Authors contributed equally

Corresponding Author:

Patrick Long, PhD

Real World Solutions, IQVIA

1 IMS Drive

Plymouth Meeting, PA, United States, 19462

Email: patrick.long@us.imshealth.com

Tel: 1-215-317-6802

WC: 2,999

**ABSTRACT**

**Objectives** To develop and evaluate machine learning models to detect suspected undiagnosed nonalcoholic steatohepatitis (NASH) patients for diagnostic screening and clinical management.

**Methods** In this retrospective observational noninterventional study using administrative medical claims data from 1,463,089 patients, gradient-boosted decision trees were trained to detect likely NASH patients from an at-risk patient population with a history of obesity, type 2 diabetes mellitus (T2DM), metabolic disorder, or nonalcoholic fatty liver (NAFL). Models were trained to detect likely NASH in all at-risk patients or in the subset without a prior NAFL diagnosis (non-NAFL at-risk patients). Models were trained and validated using retrospective medical claims data and assessed using area under precision recall and receiver operating characteristic curves (AUPRCs, AUROCs).

**Results** The 6-month incidence of NASH in claims data was 1 per 1,437 at-risk patients and 1 per 2,127 non-NAFL at-risk patients. The model trained to detect NASH in all at-risk patients had an AUPRC of 0.0107 (95% CI 0.0104 - 0.011) and an AUROC of 0.84. At 10% recall, model precision was 4.3%, which is 60x above NASH incidence. The model trained to detect NASH in non-NAFL patients had an AUPRC of 0.003 (95% CI 0.0029 - 0.0031) and an AUROC of 0.78. At 10% recall, model precision was 1%, which is 20x above NASH incidence.

**Conclusion** The low incidence of NASH in medical claims data corroborates the pattern of NASH underdiagnosis in clinical practice. Claims-based machine learning could facilitate the detection of probable NASH patients for diagnostic testing and disease management.

**SUMMARY**

What is already known on this topic

- Nonalcoholic steatohepatitis (NASH) is difficult to detect without an invasive liver biopsy and is underdiagnosed despite the risk of progression to cirrhosis.
- Machine learning (ML) models trained on real-world data have shown promise in detecting rare and underdiagnosed diseases such as NASH.

What this study adds

- This study extends the existing literature on ML applications in healthcare by incorporating high coverage medical claims data as a scalable strategy for detecting likely NASH patients.

How this study might affect research, practice or policy

- This study may increase awareness of NASH underdiagnosis and underreporting in clinical practice.
- This study may serve as a basis for future research aimed at validating ML models to support NASH diagnosis in the clinical setting.

**BACKGROUND**

Nonalcoholic fatty liver disease (NAFLD) is an umbrella term that describes two subtypes of liver disease: nonalcoholic fatty liver (NAFL) and nonalcoholic steatohepatitis (NASH).[1,2] NAFL is characterized by fat accumulation (steatosis) in the liver without significant inflammation. NASH is a more severe form of NAFLD and is characterized by steatosis with inflammation and fibrosis, which can progress to cirrhosis.[1] The prevalence of NAFLD in the United States (US) is estimated to be 24-26% of adults, of whom an estimated 20-30% have NASH.[3]

The transition from simple hepatic steatosis to NASH is a crucial point in the development of severe liver disease, putting patients at higher risk for fibrosis and progression to chronic liver disease.[4] Nevertheless, NASH is often underdiagnosed in clinical practice.[5-7] This may be due to several factors. First, there is a lack of clear patient symptoms and reliable biomarkers to help identify NASH,[8,9] and there are no universal routine screening standards.[10] Second, liver biopsy is the gold standard for NASH diagnosis but is costly, invasive, complicated by sampling errors, and requires a specialist to perform.[4,11] Finally, despite ongoing clinical trials, there are currently no approved pharmacological treatments for NASH outside of India.[12] Thus, detection of NASH remains a challenge and reliable diagnostic tools, including minimal or even non-invasive techniques, are warranted.

Machine learning (ML) with real-world data may help address the underdiagnosis of common and rare diseases. We recently demonstrated the application of ML in a retrospective case-control cohort study based on a US claims database to identify undiagnosed hepatitis C virus patients.[13] For the detection of NASH, studies have yielded encouraging results using metabolomics,[14-17] electronic health records (EHR),[18-20] or combined clinical-claims data.[21] The use case of each approach may be influenced by the chosen data type, characteristics of the model training population, or the targeted application to patients with documented NAFL.

Continuing to build upon these efforts will further enable ML approaches to facilitate NASH detection.

This study examined supervised ML using medical claims as a noninvasive strategy to identify likely NASH patients who might benefit from appropriate clinical follow-up such as monitoring or diagnostic screening. We used a retrospective rolling cross-sectional study design[22] by taking multiple snapshots of patient prescription and medical claim histories to emulate patient data during real-world deployment while providing examples of NASH patients for model training at different points in the patient journey prior to diagnosis. We also evaluated both knowledge-driven ("hypothesis-driven") and automated data-driven strategies in developing clinical predictors for NASH detection.

**METHODS**

**Data Sources**

Data were extracted from IQVIA's proprietary US longitudinal prescription (LRx) and non-adjudicated medical claims (Dx) databases.[13] LRx receives nearly 4 billion US prescription claims annually with coverage of 70% to 90% of dispensed prescriptions from retail, mail order, and long-term care channels. Dx receives over 1.35 billion US medical claims annually and covers approximately 70% of American Medical Association physicians. Dx data is derived from office-based individual professionals, ambulatory, general healthcare sites, hospitals, skilled nursing facilities and home health sites and includes patient-level diagnostic and procedural information.

All data were de-identified using an automated de-identification engine prior to being accessed by IQVIA. Both LRx and Dx datasets (hereafter referred to as LRxDx) are linked using anonymous patient tokens that support IQVIA dataset interoperability and permits longitudinal linkage across patient histories. This anonymization process is certified as Health Insurance Portability and Accountability Act (HIPAA) compliant and Institutional Review Board exempt. IQVIA holds all requisite titles, licenses and/or rights to license this Protected Health Information (PHI) for use in accordance with applicable agreements. LRxDx data spanning the study period from October 1, 2015 to June 30, 2020 were used for cohort selection and predictive feature engineering. Dx data between January 1, 2010 and October 1, 2015 were used only to exclude patients with an existing ICD-9 NASH diagnosis before study initiation. This study was conducted using the Transparent Reporting of a Multivariable Prediction Model for Individual Prognosis or Diagnosis (TRIPOD) reporting guidelines.

**Patient Selection**

An overview of the patient populations used for cohort identification is shown in Figure 1A. Patients with a history of obesity, type 2 diabetes mellitus (T2DM), metabolic disorder or NAFL were selected to include individuals who also had precursors and risk factors for NASH.[23] This has the effect of enriching the patient population to ensure that an algorithm learns a pre-diagnosis footprint specific to NASH rather than simply distinguishing between healthy and sick patients or between patients with highly dissimilar symptomologies. This cohort was stratified to remove patients with claims for liver cancer, liver failure or alcohol-related liver disease, and other liver complications that might disqualify patients from clinical intervention for NASH. Additional eligibility criteria were presence in both LRxDx for at least 24 months, recorded sex, and age from 18 to 85 years at the time of model prediction. See Supplementary Table 1 for cohort stratification criteria.

**Rolling Cross-Section Study Design**

To develop algorithms to detect NASH at different points of the pre-diagnosis patient journey (e.g., 1 month pre-diagnosis, 3 months pre-diagnosis), we divided patient claims history into a rolling series of time-bounded cross-sections (i.e., rolling cross-sections, RCS). The application of this approach to disease detection is discussed in more detail elsewhere.[22] Briefly, we divided longitudinal patient claims data over the study period into 24-month lookback periods followed by a 6-month outcome window, each shifted by 3-month increments (Figure 1B). The lookback period was used to apply stratification criteria and extract data for feature engineering. The most recent date in the lookback period (the index date) represents the starting timepoint of model prediction.

Probable NASH patients were labeled using the earliest occurrence of two criteria:
- A NASH diagnosis in the outcome window.
- A diagnosis for nonalcohol related liver fibrosis, sclerosis, or cirrhosis during the outcome window with a NAFL diagnosis within the preceding 24 or subsequent 6-months. These patients were presumed undiagnosed or unrecorded NASH patients as this clustering of diagnoses is indicative of NASH[1] and are therefore referred to as NASH proxy patients. These patients had no other liver diagnosis for their cirrhosis.

See Supplementary Table 2 for NASH labeling criteria.

A total of 152,476 patients met the selection criteria for NASH patient labeling. Patients who met the selection criteria during the lookback period without evidence of NASH during or before the cross-section outcome window formed an at-risk control patient pool. A total of 54,976,837 at-risk control patients were initially identified and then randomly downsampled to a ratio of 1 NASH patient per 5 at-risk control patients per cross-section to facilitate model training resulting in 1,312,351 at-risk patients for adversarial training.

Two modeling strategies were undertaken for NASH detection. The first (Figure 1C, NAFL Inclusive Modeling) sought to detect NASH among all at-risk patients to maximize clinical impact. We also hypothesized that patients with documented NAFL might already be suspected of NASH and that this might limit the clinical utility of algorithm-based NASH screening. We therefore investigated a second modeling approach (Figure 1C, Non-NAFL Modeling) by excluding patient cross-sections with a NAFL diagnosis claim during the lookback period.

**Feature Engineering**

We used two methods for feature engineering. The first, a knowledge-driven (KD) approach applied domain expertise to curate clinical codes into medical concepts associated with NASH risk such as relevant comorbidities, symptoms, procedures, and treatments. The second, a data-driven (DD) approach, extracted clinical codes that were present in the NASH or at-risk control patient cross-section lookbacks. Codes of each type, e.g., diagnoses or procedures, were then selected based on the largest absolute difference in prevalence between NASH and at-risk control patients with the motivation that these codes should be discriminatory predictors. DD feature identification was performed only with training cross-sections 1-6 to avoid data leakage from future cross-sections used for model validation. Patient demographics (age and sex) were included as model predictors to complement KD/DD feature sets and were included in all modeling scenarios.

Date differences and frequencies for claim and specialty visits were calculated over each patient lookback period. Date differences were calculated as the number of days between the first and last claim for a given feature relative to the index date and the number of days between the first and last claim. Missing data were represented as zero for frequency features and as null for date difference features since null values are handled inherently by the ensemble algorithm used. See Supplementary Tables 3-5 for clinical concepts used to derive model predictors.

**Model Selection**

Gradient boosted trees[24] using the xgboost package[25] were trained to discriminate between NASH and at-risk control patients. XGBoost was selected based on its previous success in benchmarking against other disease detection algorithms using claims data,[13,26] its suitability over deep learning for tabular data,[27] and its compatibility with sparse claims datasets and scalability.[25] Training cross-sections 1-6 were used for recursive feature elimination for feature selection (Supplementary Figure 1) and for hyperparameter optimization using grid search (Supplementary Table 6).

**Model Evaluation**

Models were evaluated using the area under the precision recall curve (AUPRC).[28] To compensate for random downsampling of the at-risk control cohort, precision was scaled to the 6-month incidence of NASH in the at-risk patient population before downsampling, which was 1 per 1,437 and 1 per 2,127 patients for the NAFL inclusive and non-NAFL cohorts, respectively. Such scaling ensures that the number of false positives used to calculate precision is not underestimated and accurately reflects the incidence of NASH captured in claims data. Ninety-five percent confidence intervals for model precision were approximated by treating each recall decile as a binomial distribution with n patients and a Bernoulli trial success probability of p. We then determined the uncertainty of p using a beta distribution with true and false positive predictions. The receiver operating characteristic curve and the corresponding area under the curve (AUROC) are given for reference. Feature importance was examined using SHAP (SHapley Additive exPlanations).[29]

Algorithms were compared to screenings for NASH using evidence of NAFL or T2DM in the last 2 years of a patient's claims history (see Supplementary Table 7 for qualifying clinical codes). Precision and recall of NAFL or T2DM screening were measured in NAFL inclusive and non-NAFL holdout sets. Model precision was then measured at the corresponding recall of each screening. The non-NAFL model was compared only to screening with T2DM.

**RESULTS**

**Study Cohort**

The NAFL inclusive and non-NAFL modeling cohorts displayed similar clinical profiles. T2DM was more common in NASH patients (67.1% NASH and 56.3% at-risk and 70.4% NASH and 56.4% at-risk for the NAFL-inclusive and non-NAFL patient cohorts, respectively). Obesity was common in both NASH and non-NASH patients regardless of NAFL history (59.8% vs. 59.3%). NAFL was 10-fold more common in NASH patients compared with at-risk controls (Table 1).

|  | NAFL Inclusive Modeling | | Non-NAFL Modeling | |
|---|---|---|---|---|
|  | NASH | At-Risk Controls | NASH | At-Risk Controls |
| Patients (n) | 152,476 | 1,312,351 | 104,219 | 1,265,649 |
| Patient Cross Sections (n) | 265,785 | 1,328,897 | 172,423 | 1,281,376 |
| **Demographics** | | | | |
| Age in years, mean (SD) | 57.3 (13.4) | 55.8 (15.7) | 57.6 (13.4) | 55.8 (15.9) |
| Sex (female) (%) | 59.8 | 59.4 | 59.0 | 58.0 |
| **NASH Identification Criteria (%)** | | | | |
| NASH ICD-10 | 70.3 | 0 | 76.7 | 0 |
| NASH Proxy | 29.7 | 0 | 23.3 | 0 |
| **Inclusion Criteria (%)** | | | | |
| Type 2 Diabetes | 67.1 | 56.3 | 70.4 | 56.4 |
| Obesity | 58.9 | 59.3 | 58.9 | 59.3 |
| Metabolic Syndrome | 2.5 | 1.3 | 2.4 | 1.3 |
| NAFL | 35.1 | 3.6 | 0 | 0 |
| **Comorbidities (%)** | | | | |
| Hypertension | 64.0 | 49.4 | 62.6 | 49.0 |
| Morbid (Severe) Obesity | 17.4 | 10.0 | 17.0 | 9.7 |
| Abnormal Results of Liver Function Studies | 14.0 | 1.5 | 9.2 | 1.2 |
| Abnormal Levels of Other Serum Enzymes | 12.3 | 1.8 | 8.2 | 1.5 |
| **Procedures (%)** | | | | |
| Liver Biopsy | 1.0 | <1.0 | <1.0 | <1.0 |
| Liver Panel | 14.7 | 5.2 | 10.5 | 4.9 |
| Abdominal Ultrasound | 29.1 | 5.7 | 14.6 | 4.1 |
| Comprehensive Metabolic Panel | 9.9 | 7.3 | 9.2 | 7.2 |
| **Specialty Visits (%)** | | | | |
| Gastroenterology | 30.9 | 12.7 | 22.4 | 12.0 |
| Endocrinology | 10.9 | 6.1 | 10.0 | 6.0 |
| Cardiology | 30.4 | 20.6 | 28.2 | 20.2 |

**Table 1: Cohort Statistics.** Diagnoses, procedures and physician specialty visit information is derived from medical and prescription claims data captured during each patient cross-section lookback period.

**Model Performance**

NAFL inclusive modeling detected probable NASH patients in the at-risk holdout population with 0.0107 AUPRC (95% CI 0.0104 – 0.011) and 0.84 AUROC (Figure 2A and 2B). At 10% recall, the model detected probable NASH patients with 4.3% precision (Figure 2A). This represents a 60-fold improvement over the 6-month incidence of NASH i.e., if at-risk patients were randomly screened for NASH (Figure 3A). NASH patients labeled with a NASH diagnosis and NASH proxy criteria were detected with an AUPRC of 0.0059 and 0.0051, respectively (Supplementary Figure 3).

NAFL is a precursor of NASH; however, only 35% of NASH patients received a NAFL diagnosis claim during their 24-month look-back period (Table 1). Therefore, we developed a second model to detect NASH in the non-NAFL patient cohort. The non-NAFL model identified probable NASH patients with 0.003 AUPRC (95% CI 0.0029 – 0.0031) and 0.78 AUROC (Figure 2C and 2D). At 10% recall, this model detected probable NASH patients with 1% precision, a 20-fold improvement over the incidence of NASH in this cohort (Figure 3B). A comparable number of NASH patients without NAFL were detected at approximately 30% recall by the NAFL inclusive model.

For both NAFL inclusive and non-NAFL cohorts, model precision and recall in the holdout set (i.e., train on cross-sections 1-6, test on 8) were comparable to that of the prospective validation set (i.e., train on cross-sections 1-8, test on 10) (Figure 2A). In addition, models trained on cross-sections 1-6 and then tested on either the holdout or prospective validation set showed comparable performance (Supplementary Figure 2), indicating model stability over a 6-month period.

ML surpassed the precision of NASH detection when screening at-risk patients with NAFL or T2DM claims and is presented as a fold improvement over the baseline incidence of NASH in claims data. In the NAFL inclusive cohort, NAFL screening detected 36% of likely NASH patients with 9.9-fold precision (Figure 3C), whereas the NAFL inclusive model detected the same number of NASH patients (i.e., equivalent recall) with 16.1-fold precision (Figure 3C). Screening with T2DM detected 66% of likely NASH patients with 1.2-fold precision versus 4.5-fold precision with the NAFL inclusive model (Figure 3D). In the non-NAFL cohort, T2DM screening detected 70% of likely NASH patients with 1.3-fold precision, whereas the non-NAFL model improved precision by 2.3-fold for the same recall (Figure 3E).

To evaluate the relative effectiveness of each feature engineering strategy, we compared the performance of models optimized with a mixture of KD and DD features to those optimized with one of the two feature types. Models developed with DD features performed slightly better than models developed with KD features. However, detection of NASH was maximized using the combination of KD and DD features (Figure 2).

**Model Interpretation**

To gain insight into which clinical factors drive algorithmic detection of NASH, we examined feature importance using SHAP (Figure 4). As attributes (e.g., claim frequency and claim recency) may be correlated within a single feature, we ranked the top features by taking the cumulative sum of the absolute SHAP values for attributes within each feature. For the NAFL inclusive model, a prior NAFL diagnosis was the dominant clinical predictor accounting for 13% of the total contribution (Figure 4A). In contrast, the top features were more evenly distributed in the non-NAFL model (Figure 4B). Although trained independently, KD and DD models relied on similar clinical event types including comorbidities (T2DM), laboratory findings (abnormal liver function studies and abnormal serum enzyme levels) and diagnostic procedures (abdominal ultrasonography) (Figure 4C-F).

**DISCUSSION**

This study provides encouraging results for the use of medical claims data and ML to detect likely NASH patients from large at-risk patient populations. Although there are no universally accepted routine screening standards for NASH,[10] both NAFL and T2DM are well-recognized risk factors.[2] Nonetheless, claims-based algorithms outperformed NASH screening using NAFL or T2DM history alone. In addition, algorithms detected probable NASH in at-risk patients without documented NAFL, potentially representing an overlooked NASH risk group. Approaches such as this may support more targeted screening of prevalent and underdiagnosed diseases and may be particularly valuable when diagnosis requires invasive or costly procedures or when clinical risk factors that could be used to screen patients are imprecise.

We investigated two methods that may broadly inform ML healthcare applications. First, we evaluated an automated DD approach to feature engineering for algorithmic disease detection. Model performance was greatest when KD and DD features were combined, suggesting that the two feature engineering methods make complementary contributions by integrating clinical and epidemiological knowledge with an empirically oriented process of scientific inquiry. Such automated approaches may improve knowledge discovery in real-world data while reducing the burden on clinical domain experts. Second, our RCS method can facilitate the creation of cohorts using longitudinal health data and provide opportunities to monitor healthcare market dynamics that may impact model performance.

These models represent claims-based screening tools to facilitate the identification of likely NASH patients who may benefit from clinical follow-up (e.g., via FibroScan) and as proof of concepts for further clinical validation. Potential users of these models include providers or payers who wish to implement high volume screening for suspected NASH in an at-risk patient population. While the NAFL inclusive model is suited for broad NASH detection, the non-NAFL model may be appropriate for screening NASH patients for whom NAFL status is not well documented or a reliable cause of medical care.

There are several limitations in this study. First, model precision is likely underestimated due to NASH underdiagnosis and underreporting,[5-7] which may inflate the false positive rate in model evaluation. Changes in clinical practice that facilitate NASH diagnosis should close the gap between observed incidence in claims and epidemiological estimates while also enabling the development of more powerful claims-based models. Second, our NASH labeling criteria do not guarantee NASH, which requires histological confirmation with a liver biopsy. The low percentage of patient cross-sections with a liver biopsy claim in this study may be due in part to limited coverage of this procedure in this dataset. In addition, liver biopsy may be not be performed in all cases, as a 2014 survey found that less than 25% of gastroenterologists and hepatologists routinely perform a biopsy to confirm NASH diagnosis.[6] Clinical validation of these models would need to be performed on liver biopsy-confirmed NASH patients and ideally assessed using multiple distinct medical claims datasets. Third, models were trained to detect probable NASH patients regardless of NASH stage. Additional data types may enable stage specific NASH labeling for more targeted clinical interventions. Fourth, this study used a 6-month outcome window for NASH detection, which was chosen to allow indexing on more recent claims data. However, progressive diseases such as NASH may also benefit from longer prediction horizons to enable earlier detection. Finally, the robustness of our feature engineering strategy should be determined in subsequent sensitivity analyses using a broader range of techniques such as cost-sensitive or semi-supervised learning to address class imbalances[30] as well as alternative ML algorithms and clinical targets.

**Conclusions**

In this study, we investigated claims-based ML as a noninvasive and scalable approach to stratify probable NASH patients from an at-risk population for clinical follow-up. We also demonstrated automated DD feature engineering and an RCS study design in the development disease detection algorithms. Models from this study or derivatives thereof could support more precise screening for NASH and help connect patients with available and emerging therapies.

**List of abbreviations**

AUPRC: Area under precision recall curve; AUROC: Area under receiver operating characteristic curve; DD: Data-driven; Dx: US longitudinal non-adjudicated medical claims database; EHR: Electronic health record; HIPAA: Health Insurance Portability and Accountability Act; KD: Knowledge-driven; LRx: US longitudinal non-adjudicated prescription claims database; ML: Machine learning; NAFL: Nonalcoholic fatty liver; NAFLD: Nonalcoholic fatty liver disease; NASH: Nonalcoholic steatohepatitis; RCS: Rolling cross-sections; SHAP: SHapley Additive exPlanations; T2DM: Type 2 diabetes mellitus; TRIPOD: Transparent Reporting of a Multivariable Prediction Model for Individual Prognosis or Diagnosis; US: United States; XGBoost: Extreme gradient boosting algorithm.

**DECLARATIONS**

**Ethics approval and consent to participate** The claims data used in this study were previously collected and statistically deidentified and are compliant with the deidentification conditions set forth in Sections 164.514 (a)-(b)1ii of the Health Insurance Portability and Accountability Act of 1996 Privacy Rule. No direct patient contact or primary collection of individual human patient data occurred. Study results were in tabular form and aggregate analyses, which omitted patient identification information. As such, the study did not require institutional review board review and approval or patient informed consent.

**Consent for publication** Not applicable.

**Availability of data and materials** All claims data used for this study belong to IQVIA.

**Competing interests** All authors are employees of IQVIA.

**Funding** No external funding was received to support this study.

**Contributors** All authors contributed to the conception and design of the study and the interpretation of the results. PL led the drafting of the manuscript. Data collection and analysis were performed by OY, PL, and BH. All authors contributed to the preparation and submission of the final manuscript.

**Acknowledgements** We are very grateful to Paranjoy Saharia, HEOR Scientific Services, IQVIA, and Rehan Ali, PhD and Benjamin North, PhD, Real World Solutions, IQVIA, for manuscript support.

**FIGURES**

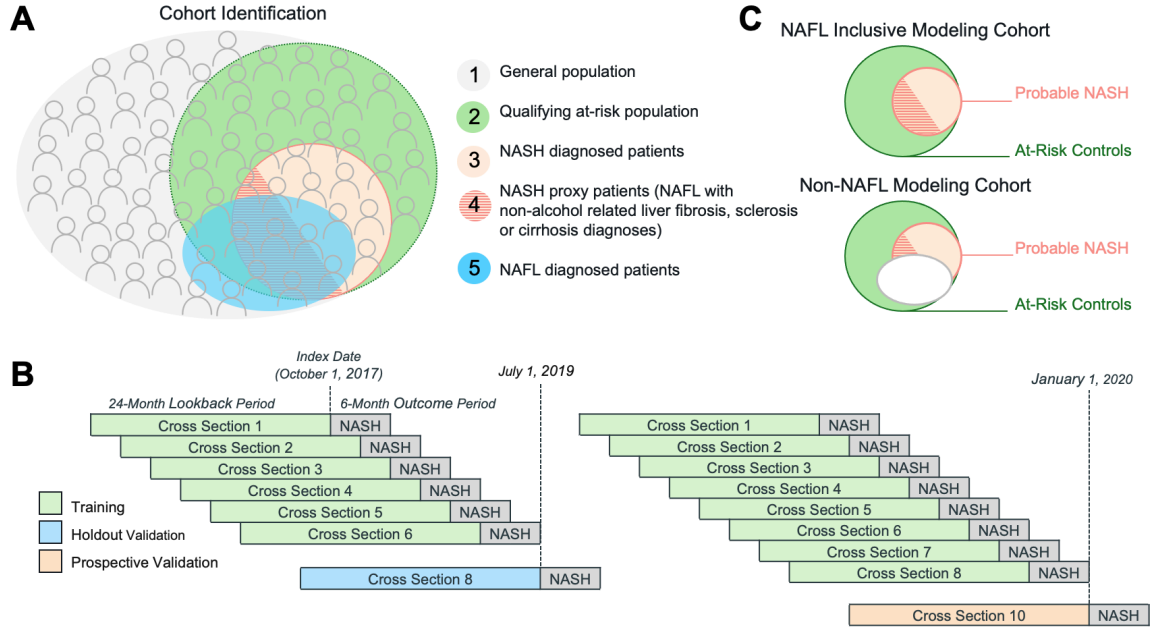

**Figure 1: Cohort Selection and Model Development.** A) Patients were identified from the general population (1) who met study stratification criteria (2). A subset of eligible patients was classified as likely to have NASH as evidenced by either a NASH diagnosis within 6-months after model prediction (3) or a diagnosis for nonalcohol related liver fibrosis, sclerosis, or cirrhosis within 6-months after model prediction and in proximity to a NAFL diagnosis (4). Eligible patients with no evidence of NASH were used as adversarial training examples, i.e., patients with an overlap in symptoms, treatments and timing of resource utilization but who were not NASH patients. Patients with an existing NAFL claim in the previous 24 months overlapped with the probable NASH patient population (5). Multiple time-bound cross-section were derived from the at-risk patient population (B). Cross-section 8 was reserved as a holdout set for model validation since it was sufficiently offset to prevent temporal overlap with the training set outcome window. Simulated model deployment using a prospective validation (scoring) set was performed by training on cross-sections 1-6 or on cross-sections 1-8 and using cross-section 10 as the test set. The NAFL inclusive and non-NAFL modeling cohorts used for model development (C).

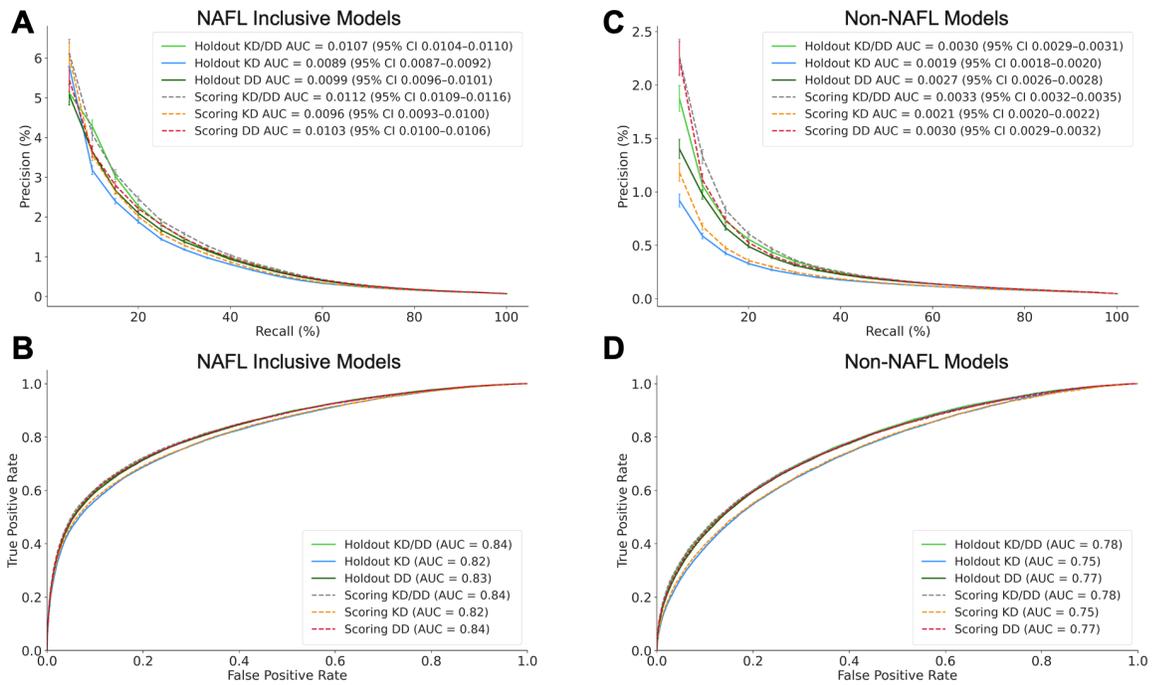

**Figure 2: Model Performance**. PR and ROC curves for model performance in detecting NASH in the NAFL inclusive (A and B) and non-NAFL (C and D) holdout and scoring validation sets. Precision is scaled to the 6-month NASH incidence observed in claims data.

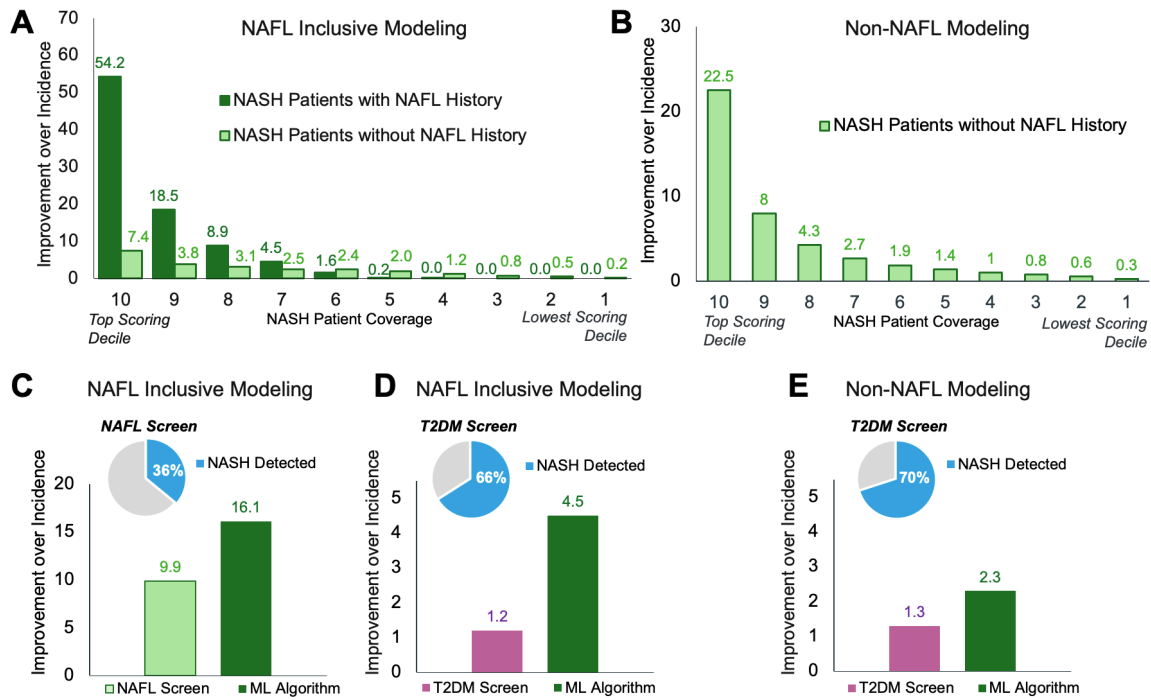

**Figure 3: Model Benchmarking.** The fold improvement in model precision over the 6-month incidence of NASH observed in medical claims data and proportions within recall deciles of predicted NASH patients with and without a NAFL diagnosis during the lookback period (A and B). Benchmark comparisons between NASH detection by ML algorithms and NASH screening using NAFL (C) or T2DM (D and E). The fold improvement over precision is calculated as the precision within each recall quantile divided by NASH incidence.

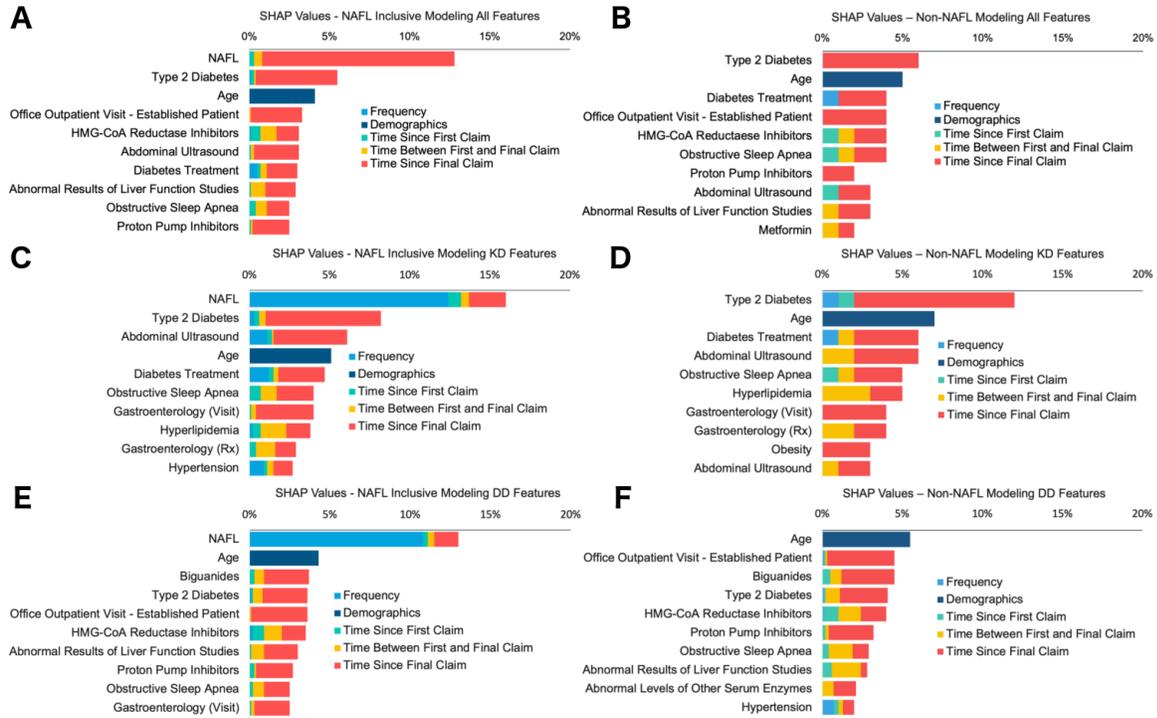

**Figure 4: Feature Importance.** SHAP values for top model features and the contribution of feature attributes i.e., claim frequency, time from first and final claim relative to the cross-section index date, time between the first and final claim occurrence, and patient demographics for NAFL inclusive and non-NAFL combined KD/DD feature models (A-B), KD features models (C-D) and DD features models (E-F). SHAP values are expressed as a percentage of the total mean absolute SHAP values for models deployed on the holdout validation set.